\pdfoutput=1

\documentclass[11pt]{article}

\usepackage[final]{acl}

\usepackage{times}
\usepackage{latexsym}
\usepackage{authblk}

\usepackage[T1]{fontenc}

\usepackage[utf8]{inputenc}

\usepackage{microtype}

\usepackage{inconsolata}
\usepackage{multicol}
\usepackage{todonotes}
\usepackage{adjustbox}
\usepackage{amsmath}
\usepackage{arydshln}
\usepackage[skins]{tcolorbox}
\usepackage{paralist}
\usepackage[vskip=1pt,leftmargin=0.7em]{quoting}
\usepackage{xcolor}
\usepackage{xspace}
\newcommand{\genonly}{{\textsc{Gen}}\xspace}
\newcommand{\srconly}{{\textsc{Src}}\xspace}
\newcommand{\srcgen}{{\textsc{Src+Gen}}\xspace}
\newcommand{\taught}{\ensuremath{\mathcal{T}}\xspace}
\newcommand{\taughtpl}{\ensuremath{\mathcal{T}_{\text{pl}}}\xspace}

\title{Boosting Zero-Shot Crosslingual Performance \\using LLM-Based Augmentations with Effective Data Selection}

\newcommand*{\emails}{%
   \normalsize \texttt{\{barah, ashishagrawal, pjyothi\}@cse.iitb.ac.in}
}

\author{\textbf{Barah Fazili$^*$}, \textbf{Ashish Sunil Agrawal$^*$}, \textbf{Preethi Jyothi} \\ Indian Institute of Technology Bombay, India \\ \emails}

\begin{document}
\maketitle
\def\thefootnote{*}\footnotetext{These authors contributed equally to this work.}\def\thefootnote{\arabic{footnote}}
\begin{abstract}
Large language models (LLMs) are very proficient text generators. We leverage this capability of LLMs to generate task-specific data via zero-shot prompting and promote cross-lingual transfer for low-resource target languages. Given task-specific data in a source language and a teacher model trained on this data, we propose using this teacher to label LLM generations and employ a set of simple data selection strategies that use the teacher's label probabilities. Our data selection strategies help us identify a representative subset of diverse generations that help boost zero-shot accuracies while being efficient, in comparison to using all the LLM generations (without any subset selection). We also highlight other important design choices that affect cross-lingual performance such as the use of translations of source data and what labels are best to use for the LLM generations.
We observe significant performance gains across sentiment analysis and natural language inference tasks (of up to a maximum of 7.13 absolute points and 1.5 absolute points on average) across a number of target languages (Hindi, Marathi, Urdu, Swahili) and domains.\footnote{The code and data for this work is available at  \href{https://github.com/csalt-research/LLM-Based-Augmentations-with-Effective-Data-Selection}{https://github.com/LLM-Based-Augmentations}}    
\end{abstract}

\section{Introduction}

Multilingual pretrained models are a mainstay in modern NLP. To create highly-performant task-specific models across different languages, a commonly adopted paradigm is to finetune a multilingual pretrained model like XLM-R~\cite{conneau-etal-2020-unsupervised} using task-specific labeled data. In the absence of labeled data for a target language, pretrained models finetuned on task-specific data in a source language (such as English) have been shown to facilitate zero-shot crosslingual transfer~\cite{yu-joty-2021-effective,zheng-etal-2021-consistency,liu-etal-2021-preserving}.
\begin{figure}[t!]
    \centering
    \includegraphics[width=0.95\linewidth]{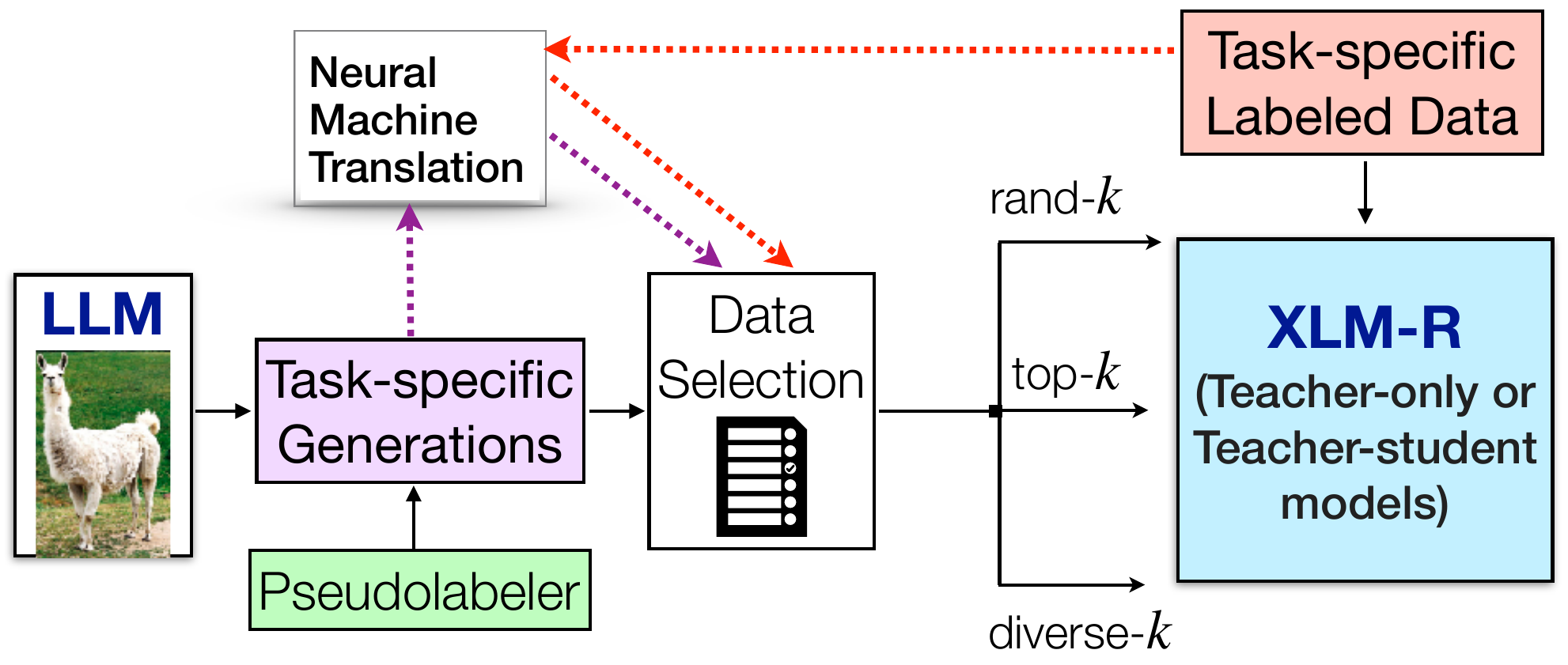}
    \caption{Overall schematic illustrating various aspects of LLM-based augmentation.}
    \label{fig:overall}
\end{figure}
Given large language models (LLMs) and their superior generation capabilities, a natural question is whether they can be used to generate synthetic task-specific data in English. To create synthetic data in a (non-English) target language, these LLM generations could be further translated into target language data using existing machine translation systems. In this work, we examine the following central question: \emph{How do we make best use of LLM generations to improve zero-shot%
\footnote{Here, by zero-shot we mean that we have no access to in-domain labeled data for a given target language. Automatic translations of English into the target language, derived from an NMT system as shown in Figure~\ref{fig:overall}, can be used during training.}
cross-lingual transfer to target languages without any labeled data?}
We stress here that we are interested in the realistic setting where task-specific data in the source language might vary in domain from the target language tasks; this setting is largely absent in zero-shot evaluations in prior work.

Our overall data augmentation pipeline is illustrated in Figure~\ref{fig:overall}. We use an open-source LLM such as Llama-2~\cite{llama2} and prompt it to generate task-specific text in English. For all target languages, we assume access to task-specific data in English that may not be in the same domain as the target-language tasks. When domain information is available for the target tasks, we add this information in the prompt to generate text that appears to be more in-domain. 


\paragraph{Pseudolabels for synthetic data.} We examine two choices to generate pseudolabels for the LLM generations: \begin{inparaenum}[(1)] \item Via prompting the LLM, to appear in the output as part of the generation. \item Via a teacher model trained on task-specific English data. \end{inparaenum}

We explore two ways in which the above-mentioned pseudolabels can be used when training models with synthetic data. We train a single model on task-specific English data augmented with generations pseudolabeled via LLM prompts. We also adopt teacher-student training where the teacher is trained on task-specific English data and the student model is trained on synthetic data with soft teacher labels (i.e. label distributions). Given label noise, we find training a student model with soft teacher labels yields significantly more accurate models compared to training with hard teacher labels.

\paragraph{Data selection.} There are two main arguments in favour of data selection:
\begin{enumerate}
\item Efficiency: A smaller subset of the generated data can be used as augmentation, thus reducing the training cost compared to using all the generated data.
\item Accuracy: Data selection helps identify training instances that are likely to aid learning more, and hence generalize better to yield overall performance improvements on downstream tasks.
\end{enumerate}

We explore different selection strategies and show that careful data selection yields stable performance improvements, unlike random selections that lead to higher variance runs and do not guarantee gains in performance. While generating synthetic data has been studied in prior work, in this work we investigate using filtering with LLM-based augmentations in a zero-shot setting.

We note here that the role of \emph{the teacher model is critical for data selection}. The teacher model gives label probabilities for every LLM generation, that is used in our data selection techniques. (With the pseudolabels derived via LLM prompts, we do not have such confidence estimates.) The utility of the teacher is mainly in identifying a suitable subset based on label probabilities. Once we identify such a subset, either LLM labels or teacher labels can be used for the instances in the subset.

\section{Methodology}
\subsection{Generation}

Consider a scenario where we have task-specific labeled data for a high-resource source language such as English (denoted as $D_{\text{en}}$). Our final downstream task is in a low-resource target language for which we have no labeled data. In this work, we experiment with two classification tasks: sentiment analysis (SA) and natural language inference (NLI). We aim to achieve improved cross-lingual transfer for these two tasks to different target languages by augmenting $D_{\text{en}}$ with (labeled) LLM generations denoted as $G_{\text{en}}$. This is motivated by recent work on boosting task performance via data augmentation techniques~\citep{strata, gal, wanli, whitehouse2023llmpowered, mrl-2023-zero}. 

$G_{\text{en}}$ is generated by prompting an LLM with a compact target domain description and the intended class label to produce class-conditioned, task-specific generations in the target domain. 
We utilize the open-source 13b llama-2-chat-hf model~\cite{llama2} for all our generations. The prompt is composed of two sub-prompts: 1) A system prompt that specifies a generic set of rules that the generator should obey, and 2) an instruction prompt that specifies more targeted instructions for generation. More details about data generation using llama-2 and the prompts for all target tasks are specified in Appendix~\ref{sec:llama2gen} and Appendix~\ref{sec:llama2prompt}.

\subsection{Pseudolabeling and Training Methods} \label{sec:train_methods}

\paragraph{Teacher-student Training (\taught).} 
A teacher model is trained on the source data ($D_{\text{en}}$) using cross-entropy loss. The teacher is used to pseudolabel the generations in $G_{\text{en}}$. A subset of $G_{\text{en}}$ is chosen via various selection techniques described in Section~\ref{dataselect}. We will refer to this subset as $D'_{\text{en}}$. A student model is trained on both $D_{\text{en}}$ and $D'_{\text{en}}$ combined, using cross-entropy loss with the gold labels in $D_{\text{en}}$ and a KL-divergence loss with the soft pseudolabels derived from the teacher model. Equation~\eqref{eq1} refers to the overall loss computed, where $y_c(x)$ is the one-hot label corresponding to each $x \in D_{\text{en}}$, $q_c$ is the student model probability for class $c$ (with temperature 1), $p_c(x)$ is the teacher model probability  for each $x \in D'_{\text{en}}$ for class $c \in C$ and $q_c^*$ is the student model probability for class $c$ (scaled by temperature value 1.5). This is the standard teacher-student paradigm, and we will refer to the trained student model as \taught in our experiments. 
\begin{align*}
\text{L$_{en}$} &=  \frac1{|D_{\text{en}}|}\sum_{x \in D_{\text{en}}} \sum_{c \in C}-y_c(x)\log{q_c(x)}+\\
&\frac1{|C||D'_{\text{en}}|}\sum_{x\in D'_{\text{en}}} \sum_{c \in C}p_c(x) \cdot \log\left(\frac{p_c(x)}{q_c^*(x)}\right)
\tag{1}
\label{eq1}
\end{align*}

Rather than using English source data and English generations, we can also adopt the translate-train setting~\cite{artetxe-etal-2020-translation} where $D_{\text{en}}$ and $G_{\text{en}}$ are translated to the target language using an off-the-shelf neural machine translation system to yield $D_{\text{tg}}$ and $G_{\text{tg}}$, respectively. The rest of the above-mentioned teacher-student training pipeline stays the same, except with using translated data everywhere. 

\paragraph{Teacher-driven Training with Prompt Labels (\taughtpl).} Instead of using a teacher model to pseudolabel the generations in $G_{\text{en}}$/$G_{\text{tg}}$, we use the teacher's label probabilities for data selection (detailed in Section~\ref{dataselect}) after which we label the data using the labels in the LLM prompts that we use for class-conditional generation. A single model is trained using cross-entropy loss on both source data in $D_{\text{en}}$/$D_{\text{tg}}$ and prompt-labeled data sampled from $G_{\text{en}}$/$G_{\text{tg}}$. The main difference from teacher-student training is the use of hard prompt labels for the sampled generations with a cross-entropy loss instead of soft pseudolabels from a teacher model with a KL-divergence loss. Here, we first utilize the teacher for data selection and subsequently use the LLM prompt labels for the generations. This model will henceforth be referred to as \taughtpl. Similar to \taught, even with \taughtpl, we can adopt the translate-train setting and use translated source data and LLM generations.

\subsection{Data Selection Strategies}
\label{dataselect}

Around 130K instances are generated for each target task from which a small subset is sampled using various data selection techniques described below. 
In all experiments, we uniformly sample across positive, negative, and neutral class labels for sentiment analysis (and entailment, contradiction, and neutral class labels for NLI) by choosing 2500 instances from the full set of instances for each class to create $D'_{\text{en}}$/$D'_{\text{tg}}$.

\begin{itemize}
\item \textbf{rand-k}: We select a random subset of 2500 instances from the data generated for each class in $G_{\text{en}}$/$G_{\text{tg}}$.
\item \textbf{top-k}: Instances specific to each class in $G_{\text{en}}$/$G_{\text{tg}}$ are sorted in descending order using the teacher model's predicted probability for that class. The top-k ($k=2500$) instances from each class are then selected.
\item \textbf{div-k}: We aim to select a diverse set of sentences from each target class using div-k. The sentences belonging to each class (based on teacher labels) are encoded using LABSE sentence embeddings \citep{labse}. The embeddings for each class are then clustered using NLTK's Kmeans clustering algorithm\footnote{https://tedboy.github.io/nlps/generated/nltk.cluster.html}. We create 25 clusters for each class and select the top 100 instances using the probabilities assigned by the teacher model (as in top-k) per cluster to get a total of 2500 instances per class. With this simple cluster-then-topk technique, we hope to identify samples that offer good coverage and capture the diversity of samples within each class. 
\item \textbf{amb-k and easy-k}: We design two additional selection techniques amb-k and easy-k by drawing inspiration from prior work on data cartography~\citep{datacart} where data points are characterized as ambiguous, easy or hard based on the training dynamics across epochs. We first compute predicted probabilities for each class for each instance across checkpoints of the teacher model saved after each training epoch. Next, we compute the mean and standard deviation across probabilities for each instance across training epochs. For each class, instances with the top-k ($k = 2500$) mean and standard deviation values are chosen as easy-k and amb-k, respectively. High standard deviation values signify larger variability in predictions across training; these instances are characterized as ambiguous examples that the model is unsure about. High mean values signify higher confidence in predictions; these instances are characterized as easy examples that the model is confident about. This selection technique is expensive in having to maintain checkpoints for all training epochs; we evaluate this only for NLI.
\end{itemize}

\section{Experimental Setup}

\subsection{Datasets} \label{sec:datainfo}

Source data refers to labeled task-specific data in English, while target data refers to evaluation sets in the target languages for which there is no labeled data. Unless specified otherwise, we choose source data to be from a different domain compared to the target data. This is different from most prior work in zero-shot evaluations where the source data is typically chosen to be consistent in the domain to the target tasks \citep{whitehouse2023llmpowered, datascarce, selfimprovepre, strata}. We assume a more realistic setting where the source and target domains can be mismatched.
 
\paragraph{Source data.} We use SST5 \citep{sst5-paper} and SNLI \citep{snlipaper} datasets for sentiment analysis (SA) and natural language inference (NLI), respectively. SST5 is a sentiment classification dataset featuring five distinct labels: negative, very negative, positive, very positive, and neutral, that we collapse into three labels: positive, negative and neutral to match the target tasks. Similar to~\citep{datascarce}, we consider a random subset of the SNLI train set (15K training sentences, 5K per class) to simulate a low-resource setting and for quicker experimental turnaround. 

\paragraph{Target data.} Our target SA tasks include 
Marathi Sentiment \citep{marsentiment-data}, GLUECoS Hindi-English code-switched Sentiment \citep{gluecos}, and Hindi Product Reviews \citep{hiproduct-data}. For NLI, we evaluate on Hindi, Urdu, and Swahili from the XNLI \citep{xnli-data} corpus; these are some of the the least-represented XNLI languages. 
Appendix~\ref{sec:datasets} provides more details about the source and target tasks. Appendix \ref{sec:codemixed} shows how we generate code-mixed data for the translate-train setting of the GLUECos task.

\subsection{Model and Training details}
For all our experiments, we use the xlm-roberta-large model \citep{xlmr} for modeling both the student and teacher. It is a 561M parameter model.\footnote{https://huggingface.co/xlm-roberta-large} Our choice of XLM-R for classification tasks was motivated by recent work on cross-lingual classification \citep{artetxe-etal-2023-revisiting} that uses only XLM-R for all its evaluations. There is also prior work \citep{zhang-etal-2023-multilingual} that shows that compared to much larger multilingual LMs like BLOOMZ, etc., fine-tuned models of smaller scale like XLM-R are at par or superior on many cross-lingual classification tasks for low-resource languages. Both the student and teacher models are trained for 15 epochs, with a learning rate of 5e-6, AdamW as the optimizer, batch size of 32, and gradient accumulation step size of 4. The student model uses a temperature of 1.5 for the KL-divergence loss. We use the best checkpoint model for all the evaluations, where the best checkpoint is selected based on accuracy over the source dev set. Translations are obtained using IndicTrans2~\citep{indictrans2} for all the languages except Swahili, for which we use NLLB~\citep{nllb}.

\begin{table*}[t!]
\centering
\begin{adjustbox}{max width=\textwidth}
\begin{tabular}{|p{0.10\linewidth}|p{0.04\linewidth} p{0.04\linewidth}|p{0.04\linewidth}p{0.05\linewidth}|p{0.04\linewidth}p{0.04\linewidth}|p{0.04\linewidth}p{0.05\linewidth}|p{0.04\linewidth}p{0.04\linewidth}|p{0.04\linewidth}|}
\hline 
& \multicolumn{2}{|c|}{\small MarSent} & \multicolumn{2}{|c|}{\small HinProd} & \multicolumn{2}{|c|}{\small XNLI Hi}& \multicolumn{2}{|c|}{\small XNLI Ur} &  \multicolumn{2}{|c|}{\small XNLI Sw} &
\multicolumn{1}{|c|}{\small Avg}\\
\hline
{\small Models} & \small dev & \small test & \small dev & \small test & \small dev & \small test & \small dev & \small test & \small dev & \small test & \\
\hline

{\small \genonly} & {\small 61.13} & {\small 61.64} & {\small 58.86} & {\small 59.43} & {\small 59.98} & {\small 60.66} & {\small 56.61} & {\small 57.35} & {\small 56.71} & {\small 56.23} & \small 58.86\\

{\small \srconly} & {\small 64.17} & {\small 63.56} & {\small 56.98} & {\small 57.94} & {\small 67.67} & {\small 67.72} & {\small 61.65} & {\small 61.34} & {\small 59.16} & {\small 58.54} & \small 61.87\\

{\small \srcgen} &  {\small 65.70} & {\small 65.91} &  {\small 62.11} & {\small 61.67} & {\small 69.74} & {\small 69.29} & {\small 65.28} & {\small 64.64} & {\small 63.52} & {\small 64.30} & \small 65.22 \\
\hline
\hline
{\small \taught-top-k} & {\small 65.68} & \textbf{\small 66.53} & \textbf{\small 68.65} & \textbf{\small 68.80}  & \textbf{\small 71.55} & \underline{\small 71.41} & {\small 64.40} & {\small 63.45} & {\small 63.82} & {\small 62.90} & \textbf{\small 66.72}\\
{\small \taught-rand-k} & {\small 65.42} & {\small 65.29} & {\small 62.11} & {\small 62.68} & {\small 71.05} & {\small 71.12} & {\small 63.01} & {\small 61.86} & {\small 62.11} & {\small 60.97} & \small 64.56\\
{\small \taught-div-k} & \underline{\small 65.81} & {\small 65.99} & \underline{\small 65.55} & \underline{\small 66.83} & {\small 71.26} & \textbf{\small 71.65} & {\small 65.82} & {\small 64.87} & \underline{\small 64.44} & {\small 64.10} & \small \underline{66.63} \\

\hdashline
{\small \taughtpl-top-k} & {\small \textbf{66.13}} & \underline{\small 66.02} & {\small 53.98} & {\small 56.50} & {\small 70.40} & {\small 70.39} & {\small \underline{66.27}} & {\small 64.06} & {\small 64.38} & {\small 63.78} & \small 64.19 \\
{\small \taughtpl-rand-k} &
{\small 64.25} & {\small 64.34} & {\small 58.06} & {\small 60.71} & {\small 71.05} & {\small 71.12} & {\small \textbf{66.43}} & \textbf{\small 65.18} & {\small 63.88} & \underline{\small 64.11} & \small 64.91 \\
{\small \taughtpl-div-k} & {\small 65.77} & {\small 65.45} & {\small 59.66} & {\small 60.61} & {\small \underline{71.31}} & {\small 70.88} & {\small 65.91} & \underline{\small 65.01} & \textbf{\small 65.68} & \textbf{\small 65.34} & \small 65.56 \\
\hline
\hline

{\small Delta} & {\small 0.43} & {\small 0.62} & {\small 6.54} & {\small 7.13} & {\small 1.81} & {\small 2.36} & {\small 1.15} & {\small 0.54} & {\small 2.16} & {\small 1.04} & \small 1.50\\
\hline
\end{tabular}
\end{adjustbox}

\caption{\label{maintable-1} This table shows the translate-train accuracies. The top three rows represent the baselines. The highest accuracy is shown in bold; the second highest is underlined. Delta represents the difference between the best-performing technique and the best-performing baseline.}
\end{table*}

\begin{table*}[t!]
\centering
\begin{adjustbox}{max width=\textwidth}
\begin{tabular}{|p{0.1\linewidth}|p{0.05\linewidth}| p{0.05\linewidth}|p{0.05\linewidth}|p{0.05\linewidth}|p{0.05\linewidth}|p{0.05\linewidth}|p{0.05\linewidth}|p{0.05\linewidth}|p{0.05\linewidth}|p{0.05\linewidth}|}
\hline 

\multicolumn{1}{|c|}{\small -} & \multicolumn{1}{|c|}{\small \genonly} & \multicolumn{1}{|c|}{\small \srconly}& \multicolumn{1}{|c|}{\small\srcgen} &  \multicolumn{1}{|c|}{\small \taught-top-k} &  \multicolumn{1}{|c|}{\small \taught-rand-k}  &  \multicolumn{1}{|c|}{\small \taught-div-k }   &  \multicolumn{1}{|c|}{\small \taughtpl-top-k } &  \multicolumn{1}{|c|}{\small \taughtpl-rand-k } &  \multicolumn{1}{|c|}{\small \taughtpl-div-k } &  \multicolumn{1}{|c|}{\small Delta }\\
\hline
\small GLUECoS & \small 52.00 & \small 51.67   & \small 54.33 &  \small 53.38 &  \small 53.73 & \small 55.52  &  \small 49.68 &  \small 48.50 &  \small 49.80 & \small 1.19\\
\hline
\end{tabular}
\end{adjustbox}
\caption{\label{gluecos-table} Zero-shot numbers for GLUECoS sentiment analysis }
\end{table*}

\subsection{Baselines}
\label{baselines}
\paragraph{Source only (\srconly).} Here, the model is trained on the train set of the source tasks (refer Section~\ref{sec:datainfo}). No synthetic data is used to train the model. 8,544 and 15K instances are used for SA and NLI tasks, respectively.
\paragraph{Source+Generations (\srcgen).} Here, the model is trained on a mixture of source and synthetic datasets. The synthetic dataset (7.5K) is sampled randomly from among the generations and is not selected via any data selection technique or with the help of a teacher resulting in 16K and 22.5K instances for SA and NLI tasks, respectively.

\paragraph{Generations only (\genonly).} Here, we train the model only on the synthetically generated data. The labels come from the prompts that we used for class conditional generation. For a fair comparison with \srcgen, we maintain the same total size of ~16K and 22.5K instances for SA and NLI tasks, respectively

\section{Results and Analysis}

\subsection{Main Results}

We characterize the training data used for each model along two axes: source of the task-specific text and source of labels assigned to the data. \genonly, \srconly and \srcgen in Table~\ref{maintable-1} represent the baseline models as described in Section~\ref{baselines}. Generated text used in all the baseline models is combined with the corresponding prompt labels used during generation. We observe that substituting a portion of generated instances with source instances (\srcgen) yields better performance, as anticipated, compared to using \genonly alone. Further augmenting the source data with generations (\srcgen) boosts the \srconly baseline across all evaluated tasks/languages.

The results in Table~\ref{maintable-1} are all translate-train accuracy values since they are found to largely outperform the zero-shot numbers, thus highlighting the benefits of using (machine) translations for cross-lingual evaluations (as reported in prior work~\cite{artetxe-etal-2023-revisiting}). Please refer to Appendix \ref{app:zsresults} for zero-shot results. Our reported numbers are averaged across models trained on two different random seeds.\footnote{For the Hindi SA task, due to the considerably smaller size of the evaluation sets, we trained the models using six different seeds to obtain more reliable evaluations.} Following the three rows of baseline numbers in Table~\ref{maintable-1}, we show results using models that are trained across two different levels of supervision from the \srconly baseline (acting as the teacher). \taught indicates that data selection is done using teacher pseudolabels while \taughtpl indicates that after data selection using teacher pseudolabels, for each instance, prompt labels are used for subsequent training (instead of retaining the teacher-assigned labels).
Each of the listed models is trained on data selected using various selection strategies detailed in Section~\ref{dataselect}. The \taught models are trained  with soft pseudolabels derived from the teacher while the baselines and \taughtpl models are trained with hard prompt labels. For a given strategy (top-k, rand-k, etc.), we note that the same unlabeled data subsets are used with one of the two kinds of labels (teacher soft vs. prompt hard). 

Across all tasks, we observe that data selection strategies yield consistent performance improvements over the best baseline with absolute accuracy gains of up to $7\%$ for Hindi Product SA. \taughtpl appears to do better overall i.e., prompt labels after teacher-based data selection; the teacher labels perform much better just on the Hindi Product task. We note here that unlike prior work that uses LLM-based augmentations for cross-lingual tasks~\cite{whitehouse2023llmpowered} with access to some target data, all our models are trained without \emph{any access} to real target data.

We find that the delta values in Table~\ref{maintable-1} using our data selection techniques for XNLI and Marathi SA (that have a similar number of test instances) are statistically significant at $p<0.01$ using the Wilcoxon signed rank test. Since the Hindi product review task has a significantly smaller number of test instances, we treat it separately across different random seeds and find that top-k data selection results in a statistically significant improvement (compared to \srcgen) at $p<0.05$ using the Wilcoxon signed rank test.

Other than the five target sets in Table~\ref{maintable-1}, in Table~\ref{gluecos-table} we also evaluate on a code-switched Hindi-English sentiment analysis task which is yet another challenging low-resource domain. Unlike Table~\ref{maintable-1} where translate-train was more effective, we show zero-shot scores since the presence of English words in code-switched En-Hi text is found to benefit more from \srcgen (teacher) trained on original generations in English. More details about the code-mixed text generation and the corresponding Translate-train numbers are provided in Appendix~\ref{sec:codemixed}.

\subsection{Experimental Analysis} \label{sec:analysis}
\begin{table}[t!]
\centering
\begin{adjustbox}{max width=\textwidth}
\begin{tabular}{|p{0.25\linewidth}|p{0.06\linewidth} p{0.07\linewidth}|p{0.06\linewidth}p{0.07\linewidth}|p{0.06\linewidth}p{0.07\linewidth}|}
\hline 
&\multicolumn{2}{|c|}{\small XNLI Hi}& \multicolumn{2}{|c|}{\small XNLI Ur} & \multicolumn{2}{|c|}{\small XNLI Sw}\\
\hline
 & {\small dev} & \small test & \small dev & \small test & \small dev & \small test \\
\hline
{\small \srcgen} &{\small 69.74} & {\small 69.29} & {\small 65.28} & {\small 64.64} & {\small 63.52} & \underline{\small 64.3}\\
\hline
\hline
{\small \taught-amb-k} & {\small 69.14} & {\small 70.03} & {\small 64.00} & {\small 62.49} & {\small 63.98} & {\small 63.02} \\
{\small \taught-easy-k} & {\small 69.36} & {\small 69.75} & \underline{{\small 65.47}} & \underline{{\small 64.85}} & {\small 63.17} & {\small 62.76} \\
{\small \taughtpl-amb-k} &  \textbf{\small 72.05} & \textbf{\small 71.36} & {\textbf{\small 66.91}} & {\textbf{\small 66.12}} & {\textbf{\small 65.66}} & {\textbf{\small 64.98}}\\
{\small \taughtpl-easy-k} & \underline{{\small 70.62}} & \underline{{\small 70.59}} & {\small 63.82} & {\small 62.88} & \underline{{\small 64.30}} & {{\small 63.49}} \\
\hline
{\small Delta} & {\small 2.31} & {\small 2.07} & {\small 1.63} & {\small 1.48} & {\small 2.14} & {\small 0.68} \\
\hline
\end{tabular}
\end{adjustbox}
\caption{\label{ambi-easy-xnli} Translate-train accuracies using amb-k and easy-k selection (see section~\ref{dataselect}). Delta is (best score - \srcgen) score.}
\end{table} 
\paragraph{Ambiguous/Easy Data Selection.} Table~\ref{ambi-easy-xnli} shows XNLI results of student models/prompt-based models trained on data selected using amb-k and easy-k selection techniques. 
Augmenting the source data with prompt-labeled ambiguous instances benefits the model the most. Ambiguous instances are ones that the model is most uncertain about and are likely to help the model generalize well. This is consistent with observations about ambiguous instances in prior work~\citep{datacart,wanli}.

\paragraph{Soft Labels vs. Hard Labels.} Table \ref{hardsoft-tbi} shows the translate-train accuracies of a student model (\taught) trained using teacher hard pseudo labels and soft pseudo labels. CE implies training using teacher hard labels and cross-entropy loss, KLD implies training using teacher soft labels and KL-divergence loss. Large delta values highlight the significance of using soft teacher labels instead of hard labels. If the teacher labels are noisy, soft labels help the student model generalise better to unseen data.
\begin{table}[t!]
\centering
\begin{adjustbox}{max width=\textwidth}
\begin{tabular}{|p{0.09\linewidth}|p{0.1\linewidth}|p{0.1\linewidth}|p{0.1\linewidth}|p{0.1\linewidth}|}
\hline 
&\multicolumn{2}{c}{\small Marsentiment}& \multicolumn{2}{|c|}{\small XNLI Hi}  \\
\hline
&\multicolumn{2}{c}{\small \taught-top-k} & \multicolumn{2}{|c|}{\small \taught-top-k}\\
\hline 	
 &\multicolumn{1}{c}{\small dev}& \multicolumn{1}{|c|}{\small test} & \multicolumn{1}{c}{\small dev}& \multicolumn{1}{|c|}{\small test} \\
\hline
{\small CE} & {\small 45.75} & {\small 44.43} & {\small 40.52} & {\small 40.09}\\
\hline 
{\small KLD} & \textbf{\small 65.68} & \textbf{\small 66.53} & \textbf{\small 71.55} & \textbf{\small 71.41}\\
\hline
\hline
{\small Delta} & \small 19.93 & \small 22.10 & \small 31.03 & \small 31.32 \\
\hline
\end{tabular}
\end{adjustbox}
\caption{ Compare training of \taught-top-k model with teacher-soft vs teacher-hard labels. Delta represents the difference between the accuracy (translate-train) for the student trained via soft labels and the student trained using hard labels.}
\label{hardsoft-tbi}
\end{table}

\paragraph{Effect of Varying Sizes of Augmented Data.} To study the effect of augmented data size on cross-lingual transfer, we experiment with div-k selection (\taught model) and \srcgen model for the Marathi SA task. Models are trained in the translate-train setting over varying amounts of augmented data. Table~\ref{tab:largerKs} shows that increasing $k$ leads to a decrease in accuracy. The consistent downward trend in the div-k selection technique underscores the importance of data selection; the best accuracies were obtained using div-k with $k=7500$. Also, augmenting synthetic data also results in increased training time. Determining the optimal augmentation size for each target task is left as future work.

\begin{table}[b!]
\centering
\begin{adjustbox}{max width=\textwidth}
\begin{tabular}{|p{0.2\linewidth}|p{0.1\linewidth}|p{0.1\linewidth}|p{0.1\linewidth}|p{0.1\linewidth}|}
\hline 
&\multicolumn{4}{|c|}{\small Marsentiment}\\
\hline
&\multicolumn{2}{c}{\small \taught-div-k} & \multicolumn{2}{|c|}{\small \srcgen}\\
\hline 	
 &\multicolumn{1}{c}{\small dev}& \multicolumn{1}{|c|}{\small test} & \multicolumn{1}{c}{\small dev}& \multicolumn{1}{|c|}{\small test} \\
\hline
{\small $k=2500$} & \underline{\small 65.81} & \underline{\small 65.99} & \textbf{\small 65.38} & \underline{\small 65.42}\\
\hline 
{\small $k=7500$} & \textbf{\small 66.32} & \textbf{\small 	66.57} & {\small 64.31} & {\small 64.44}\\
\hline
{\small $k=12500$} & {\small 65.28} & {\small 65.17} & {\small 63.57} & {\small 64.39}\\
\hline
{\small $k=17500$} & {\small 64.73} & {\small 64.53} & \underline{\small 65.14} & \textbf{\small 65.57}\\
\hline
{\small $k=22500$} & {\small 64.64} & {\small 64.22} & {\small 64.08} & {\small 64.79}\\
\hline
\hline
{\small Delta} & \small 1.68 & \small 2.35 & \small 1.30 & \small 0.78 \\
\hline
\end{tabular}
\end{adjustbox}
\caption{Translate-train accuracy analysis by training \taught-div-k and \srcgen models with different sizes of augmented data ($k$ is the examples augmented per class). Delta represents the difference between the best accuracy and accuracy for $k=22500$.}
\label{tab:largerKs}
\end{table}

\paragraph{Augmenting with Target Train Data.} To explore whether LLM generations boost performance even in the presence of source data that matches in domain to the target task (henceforth referred to as target training data), we train teacher models on a subset of 15K sentences from the XNLI train set. (Note that the numbers in Table~\ref{maintable-1} were obtained using SNLI as the source data.) As expected, we see significant improvements in teacher accuracies in Table~\ref{xnlitargetdomain} when using target train data. Student models trained on generations pseudolabeled with this superior teacher further boost accuracies; the best results are obtained using a combination of translate-train and div-k selection. Also, the accuracies obtained using \srcgen models are inferior to the best scores obtained using \taught models, again highlighting the importance of data selection.  
\begin{table}[t!]
\centering
\begin{tabular}{|p{0.25\linewidth}|p{0.06\linewidth}p{0.07\linewidth}|p{0.06\linewidth}p{0.07\linewidth}|p{0.06\linewidth}p{0.07\linewidth}|}
\hline 
&\multicolumn{2}{|c|}{\small XNLI Hi}& \multicolumn{2}{|c|}{\small XNLI Ur} & \multicolumn{2}{|c|}{\small XNLI Sw}\\
\hline
 & {\small dev} & \small test & \small dev & \small test & \small dev & \small test \\

\hline
{\small \srconly} & {\small 79.60} & {\small 78.10} & {\small 71.81} & {\small 70.00} & {\small 73.37} & {\small 72.28}\\
\hline

{\small \srcgen} & \small 79.32 & \underline{\small 78.47} & \small 72.13 & \small 69.96 & \small 73.34 & \underline{\small 72.33} \\
{\small \taught-top-k} & \underline{\small 79.78} & {\small 77.82} & \underline{\small 72.31} & \underline{\small 70.00} & \textbf{\small 74.24} & \textbf{\small 72.41} \\
\hline

{\small \taught-div-k} & \textbf{\small 80.44} & \textbf{\small 78.54} & \textbf{\small 72.87} & \textbf{\small 70.89} & \underline{\small 74.12} & {\small 71.70} \\
\hline
\hline
{\small Delta} & {\small 0.84} & {\small 0.44} & {\small 1.06} & {\small 0.89} & {\small 0.89} & {\small 0.13} \\
\hline
\end{tabular}
\caption{Using in-domain target train (taking a random subset from MNLI) for training the Teacher and also as the source data (translate-train accuracies).}

\label{xnlitargetdomain}
\end{table}

\paragraph{Generations Uniform across Classes.} By default, we create class-balanced augmentations by sampling 2500 instances from each class based on teacher labels. To analyse the effect of class imbalance on downstream evaluation, we augment the sentiment source data with class-imbalanced augmented data sets for the Marathi SA task. The total size of the augmented data remains constant, while the class distribution is altered by eliminating neutral sentences from 2500 to 0, thereby transitioning towards sentiment-rich augmentations. We see in Table \ref{randomsubset} that students trained on data that is uniformly distributed across classes along with the top-k selection strategy exhibits superior performance, compared to those trained on a subset of the generated data with imbalanced class proportions. This suggests that employing a class-balanced augmentation is an important consideration.
\begin{table}[t!]
\centering
\begin{adjustbox}{max width=\textwidth}
\begin{tabular}{|p{0.5\linewidth}|p{0.1\linewidth}|p{0.1\linewidth}|}
\hline 
&\multicolumn{2}{|c|}{\small \taught-top-k}  \\
\hline
&\multicolumn{1}{c}{\small dev}& \multicolumn{1}{|c|}{\small test}  \\
\hline
{\small 2500 pos, 2500 neg, 2500 neu} & \textbf{\small 65.68} & \textbf{\small 66.53}\\
\hline 
{\small 3000 pos, 3000 neg, 1500 neu} & {\small 64.28} & {\small 65.40}\\
\hline
{\small 3500 pos, 3500 neg, 500 neu} & {\small 65.13} & {\small 65.51}\\
\hline
{\small 3750 pos, 3750 neg, 0 neu} & {\small 64.98} & {\small 65.08}\\
\hline
\end{tabular}
\end{adjustbox}
\caption{Probing the effect of class imbalance on Marathi sentiment task. pos, neg, and neu indicate positive, negative, and neutral classes.  The best numbers (translate-train) in a column are highlighted.}

\label{randomsubset}
\end{table}

\paragraph{Quality of prompt labels.} To evaluate whether the LLM prompt labels are truly reflected in the filtered generated text or not, we ran a human evaluation on a set of 100 generations each for sentiment analysis and NLI. These instances were randomly selected generations from among a set of pseudolabels predicted with high probability by the teacher model for the respective tasks. The average accuracy of label alignment between annotator-provided labels and prompt-derived labels was found to be 87.88\% and 71.72\%, with Cohen's kappa coefficients of 0.752 and 0.749, respectively for the two tasks. This suggests that the prompt labels in the subset obtained after teacher-based filtering are of fairly high quality. Please refer to Appendix~\ref{annotator-details} for more details.

\begin{figure}[t!]
    \centering    \includegraphics[width=0.99\linewidth]{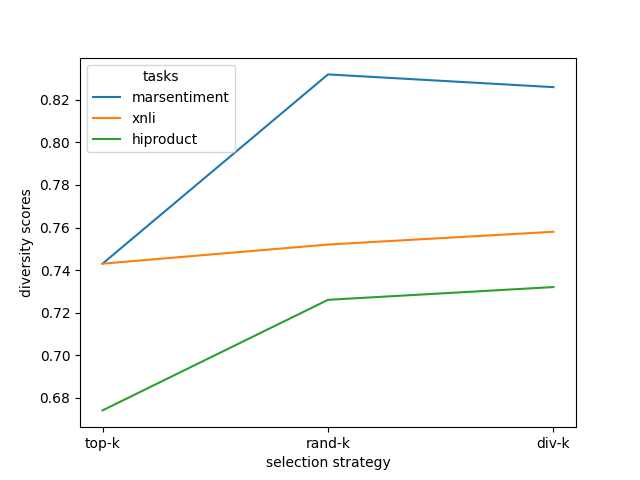}
\caption{Diversity scores of augmented data for different data selection strategies and different tasks.}
   \label{fig:diversity}
\end{figure}

\paragraph{Analyzing Diversity.} We introduce a simple metric that we call a \emph{``diversity score"} to capture the dissimilarity in text embeddings across sentences in a dataset. This is computed by encoding each instance using LABSE \citep{labse}, taking the average of the cosine distance of the LABSE embedding with every other instance in the data sample and finally taking an average of these mean distances across all data samples. To check if the data selected using the div-k selection technique is indeed diverse, we compute the diversity score for each task and data selection strategy. Figure \ref{fig:diversity} shows the trend of diversity scores. It is clear that the diversity of the 7500 sentences selected using div-k technique is greater than the diversity of the sentences selected via top-k and rand-k across all tasks.

\begin{table}[t!]
\centering
\begin{tabular}{|p{0.3\linewidth}|p{0.1\linewidth}|p{0.1\linewidth}|p{0.1\linewidth}|p{0.1\linewidth}|}
\hline 
&\multicolumn{2}{c}{\small MarSent}& \multicolumn{2}{|c|}{\small XNLI Hi}\\
\hline
&\multicolumn{1}{c}{\small dev}& \multicolumn{1}{|c|}{\small test}  &\multicolumn{1}{c}{\small dev}& \multicolumn{1}{|c|}{\small test} \\
\hline 	
{\small \srcgen-top-k}  & \small 65.38 & \small 65.42  & {\small 70.10} & {\small 70.20}  \\
\hline 	
{\small \taught-top-k (in-domain)} & \underline{\small 66.34} & \textbf{\small 66.54}  & {\small 70.82} & {\small 69.81}\\
\hline 	
{\small \taught-top-k law} & \textbf{\small 66.45} & \underline{\small 66.47}  & \textbf{\small 71.43} & \textbf{\small 70.08}  \\
\hline 	
{\small \taught-top-k medical} & {\small 66.00} & {\small 64.83}  & \underline{\small 71.26} & \underline{\small 69.97}  \\
\hline

\end{tabular}
\caption{Cross-domain experiments for Marathi sentiment and XNLI-Hi task with top-k selection. Best numbers are highlighted and the second-best are underlined.} 
\label{crossdomain}
\end{table}

\begin{table}[t!]
\centering
\begin{adjustbox}{max width=\textwidth}
\begin{tabular}{|p{0.15\linewidth}|p{0.1\linewidth}|p{0.1\linewidth}|p{0.1\linewidth}|}
\hline
& \multicolumn{1}{|c|}{\textbf{\small Pos score}} & \multicolumn{1}{|c|}{\textbf{\small Neg score}} & \multicolumn{1}{|c|}{\textbf{\small Overall score}} \\
\hline
\small Medical & \small 0.037 & \small 0.035 & \small 0.073\\
\hline
\small Law &\small 0.034 & \small 0.027  & \small 0.061\\
\hline
\small Marathi In-d & \small 0.026 & \small 0.020 & \small 0.046 \\
\hline
\end{tabular}
\end{adjustbox}
\caption{\label{cross-domain-sent-rich} Sentiment richness of Marathi sentiment in-domain and out-of-domain datasets; Pos, Neg indicates positive and negative.
}
\end{table}

\paragraph{Cross-Domain Analysis.}  Recall that the prompts to the LLM also contained domain information of the target task. To evaluate the impact of domain-specificity of the generations on zero-shot performance, we create two cross-domain datasets in the medical and law domain (unrelated to the target task domains). 

Table~\ref{crossdomain} shows zero-shot results of the student models trained on the in-domain and out-of-domain (law, medical) augmented datasets for the Marathi sentiment and XNLI Hi tasks. We observe that the student model trained on out-of-domain datasets perform comparably to the student models trained on in-domain data. This shows that models can effectively learn task information from augmented data even if it comes from domains that differ from the target task. 

To further disentangle the roles of task and domain, we made an attempt to capture sentiment richness present in the in-domain vs out-of-domain data for the Marathi SA task. We use a sentiment lexicon\footnote{We used the SentiWordNet 3.0 sentiment lexicon available at: \url{https://github.com/aesuli/SentiWordNet?tab=readme-ov-file)}.} consisting of a positive and negative sentiment score for each word along with its POS tag, word sense, etc. We simply add the corresponding scores for each word in the corpus found in the lexicon (along with matching the POS tag) and normalize the sums by the word count of the corpus. Table~\ref{cross-domain-sent-rich} shows that the computed sentiment scores for the generated out-of-domain datasets are much better than the in-domain dataset for Marathi SA. Here, overall score is calculated by summing both the positive and negative scores for each word. Hence, it is plausible that the sentiment richness in the generated out-of-domain datasets compensates for the domain mismatch, thus yielding comparable or slightly better results in Table~\ref{crossdomain}. (Tables~\ref{diffsentimentdomains}, \ref{diffnlidomains} in Appendix~\ref{sec:llama2prompt} show examples of generations from different domains.)

\section{Related Work}
Our work is closely related to \citeauthor{whitehouse2023llmpowered} that studies generations from various open-source and commercial LLMs for cross-lingual performance over reasoning tasks. However, they rely on instances from the target sets as few-shots for generations and do regular finetuning over labels derived from the class-conditional prompts.

Furthermore, our work is grounded in ideas inspired by \citeauthor{he2022generate}, incorporating self-training on unlabeled synthetic text produced by Language Models. However, \citeauthor{he2022generate} fine-tuned generators using target data, and their experiments were limited to GLUE tasks (in English). In contrast, our focus is on multilingual models, aiming for cross-lingual transfer from source data in a high-resource language across arbitrary domains in different task languages. We achieve this by self-training on zero-shot generations from LLMs without utilizing any target data during generation or training.

We draw inspiration from \citep{datacart} to design data selection techniques based on the principle of dataset cartography. ~\cite{wanli} use dataset cartography on a large NLI dataset (MNLI) to choose instances with complex reasoning patterns, and instructs GPT-3 to generate new examples with similar patterns. Automatically generated examples undergo filtering, and ultimately, human crowdworkers review, revise, and label them. In a similar vein, \citeauthor{khanuja2023demux} present language-agnostic methods to pick specific data points to be labeled from a large, unlabelled multilingual dataset. These points are chosen either by considering their distance from the target set, the uncertainty of model predictions over them, or finding a balance between minimizing distance and maximizing model uncertainty. While WANLI~\citep{wanli} only explores English generation/evaluation, both depend not only on human-in-the loop annotation of unlabeled text, but also depend on existing target data for generator finetuning \citep{khanuja2023demux} or few-shot prompting~\citep{wanli}.

\citeauthor{de-raedt-etal-2023-zero} propose in-place augmentation of data instances from high-resource languages for better out-of-distribution generalization by leveraging LLMs. However, these techniques are specifically applicable to single-text classification tasks. For tasks like NLI or QA, which involve multiple components in each instance, automatically making such edits or augmentations while preserving the intended relationships between components (and affecting labels) is not straightforward.

In contrast to \citep{paxqa}, and \citep{syntheticqa}, we make use of an open-source LLM to generate task and domain-specific synthetic data. We also explore the more realistic setting of having no access to source data that matches the target domain; this is not explored in the above two works. Similar to our work, synthetic data generation has been explored in \citep{agrawal2023qameleon, gekhman-etal-2023-trueteacher}. However, both works do not use any form of data selection for the synthetically generated data. \citep{gekhman-etal-2023-trueteacher} do not show evaluations on low-resource languages; their multilingual experiments are limited to high-resource languages such as English, French, Spanish and German. \citep{agrawal2023qameleon} show experiments in the few-shot setting, while we operate in the zero-shot setting.

\section{Conclusion}
In this work, we focus on the broader problem of boosting zero-shot cross-lingual transfer using LLM-based augmentations. We highlight the importance of using data selection strategies to select smaller subsets that result in more efficient training and improved performance on downstream target language tasks. We also compare and contrast the utility of pseudolabeling generations using labels from LLM prompts versus using a teacher model to label the generations. One of the main takeaways is that LLM generations, in conjunction with our data selection strategies, can help improve cross-lingual transfer regardless of whether task-specific source data matches the domain of the target tasks or not.

\section*{Acknowledgements}
The authors thank the anonymous reviewers for their constructive feedback  and engagement during the rebuttal that helped improve the quality of the draft. The last author would like to gratefully acknowledge a faculty grant from Google Research India supporting her research on multilingual models.

\section*{Limitations}
\begin{enumerate}
\item The generations from LLMs are sensitive to the prompts used. Although we share our custom prompts, the quality of the generated content is heavily reliant on the particular domain and task for which the data is generated creating some non-determinism.
\item Because of budget constraints, our investigations were constrained to an open-source LLM (LLAMA-2). It is possible that higher-capacity commercial LLMs could yield better performance. 
\item We explore many data selection techniques but a clear winner across all tasks/settings has not emerged. Although ambiguous selection gives best scores for XNLI, more target domains and languages should be included to study the most effective filtering techniques in general.
\item We have only experimented with generating data for classification tasks; generating data for more structured tasks like QA or commonsense reasoning tasks could pose challenges.
\item For the translate-train models, one assumes access to MT models for the target language which may not always be available. 
\end{enumerate}

\bibliography{anthology,custom}

\appendix

\section{Dataset Description}
\label{sec:datasets}
Table \ref{aux-task-data} shows the details of the source tasks. SST5 is used as a source task for all the sentiment tasks, and SNLI as a source task for the XNLI tasks. Table \ref{target-task-data} shows the details of the different target tasks. We treat both the dev and test sets of these target tasks as test sets and do not use their train sets for model training. The language "Hinglish" refers to the Hindi-English code-mixed text, the data is present in a mixed script (both Romanized and Devanagari) form. Table \ref{domain-info} shows the domain information present in the different target tasks. The domain list is not exhaustive, we made use of the domains that were easier to represent in a prompt.
\begin{table}[t!]
\centering
\begin{tabular}{|p{0.22\linewidth}|p{0.18\linewidth}|p{0.18\linewidth}|p{0.18\linewidth}|}
\hline
 \small\textbf{Task} & \small\textbf{\#Train} & \small \textbf{\#Dev} & \small \textbf{\#Test}\\
 \hline
\small SST5 & \small{8,544} & \small {1,101}  & \small {2,210}\\
\hline
\small SNLI & \small 15,000 & \small 9,842  & \small 9,824\\

\hline
\end{tabular}
\caption{\label{aux-task-data}Source task details 
}
\end{table}

\begin{table}[t!]
\centering
\begin{tabular}{|p{0.17\linewidth}|p{0.12\linewidth}|p{0.12\linewidth}|p{0.15\linewidth}|p{0.12\linewidth}|}
\hline
 \small\textbf{Task} & \small\textbf{\#Dev} & \small \textbf{\#Test} & \small \textbf{Language} & \small \textbf{Source Dataset}\\
 \hline

\small Marathi Sentiment & \small{6,000} & \small {6,750} & \small {Marathi} & \small SST5\\
\hline
\small IITP Product Review & \small{523} & \small {523} & \small {Hindi} & \small SST5\\
\hline
\small XNLI & \small{2,490} & \small {5,010} & \small {Hindi} & \small SNLI\\
\hline
\small XNLI & \small{2,490} & \small {5,010} & \small {Urdu} & \small SNLI\\
\hline
\small XNLI & \small{2,490} & \small {5,010} & \small {Swahili} & \small SNLI\\
\hline
\small GLUECos Sentiment & \small{1,260} & \small {-} & \small {Hinglish} & \small SST5\\
\hline
\end{tabular}
\caption{\label{target-task-data}Target task details 
}
\end{table}

\begin{table}[t!]
\centering
\begin{tabular}{|p{0.21\linewidth}|p{0.4\linewidth}|}
\hline
\small MarSentiment & \small Mixture of political tweets, sitcom subtitles, generic tweets and movie reviews\\
\hline
\small Hindi Product & \small Reviews about travel, movies, and various electronic gadgets\\
\hline
\small XNLI & \small Travel, fiction, government domains\\
\hline
\small GLUECos & \small Generic tweet domain\\
\hline
\end{tabular}
\caption{\label{domain-info}Target Task domains, the list is not exhaustive, we picked up domains which could be represented well in a prompt 
}
\end{table}

\begin{table*}[t!]
\centering
\begin{adjustbox}{max width=\textwidth}
\begin{tabular}{|p{0.16\linewidth}|p{0.04\linewidth}p{0.05\linewidth}|p{0.04\linewidth}p{0.04\linewidth}|p{0.04\linewidth}p{0.05\linewidth}|p{0.04\linewidth}p{0.04\linewidth}|p{0.04\linewidth}p{0.04\linewidth}|}
\hline 
&\multicolumn{2}{|c|}{\small MarSent} & \multicolumn{2}{|c|}{\small HinProd} & \multicolumn{2}{|c|}{\small XNLI Hi}& \multicolumn{2}{|c|}{\small XNLI Ur} &  \multicolumn{2}{|c|}{\small XNLI Sw}\\
\hline
{\small selection strategy} & \small dev & \small test & \small dev & \small test & \small dev & \small test & \small dev & \small test & \small dev & \small test\\
\hline
{\small \genonly} & {\small 62.28} & {\small 62.66} & {\small 60.96} & {\small 61.38} & {\small 62.89} & {\small 62.06} & {\small 57.21} & {\small 56.59} & {\small 56.02} & {\small 56.17}\\
\hline
{\small \srconly} & {\small 63.70} & {\small 62.96} & {\small 61.95} & {\small 63.48} & {\small 65.54} & {\small 64.61} & {\small 57.15} & {\small 55.43} & {\small 57.63} & {\small 55.45}\\
\hline
{\small \srcgen} &  {\small 65.38} & {\small 65.42} &  {\small 65.14} & {\small 63.16} & {\small 70.10} & {\small 70.20} & \textbf{\small 63.92} & \textbf{\small 63.81} & {\small 63.34} & \textbf{\small 63.64}\\
\hline
\hline
{\small \taught-top-k} & {\small 66.34} & \underline{\small 66.54} & \textbf{\small 66.96} & \textbf{\small 69.00} & \underline{\small 70.82} & {\small 69.81} & {\small 62.98} & {\small 62.02} & \underline{\small 63.74} & \underline{\small 62.94}\\
{\small \taught-rand-k} & {\small 66.22} & {\small 66.06} &  {\small 66.03} & {\small 67.56} & {\small 68.80} & {\small 67.68} & {\small 59.78} & {\small 58.21} & {\small 61.67} & {\small 59.49}\\
{\small \taught-div-k} & \underline{\small 66.38} & {\small 66.12} & \underline{\small 66.25} & \underline{\small 68.04} & {\small 70.38} & {\small 69.65} & {\small 62.83} & {\small 61.03} & {\small 62.69} & {\small 61.51}\\
\hdashline
{\small \taughtpl-top-k} & \textbf{\small 67.83} & \textbf{\small 68.01} & {\small 58.29} & {\small 58.67} & {\small 70.02} & {\small 69.86} & {\small 63.33} & {\small 61.73} & {\small 62.11} & {\small 61.18}\\
{\small \taughtpl-rand-k} & {\small 65.16} & {\small 65.05 } & {\small 60.77} & {\small 59.27} & \textbf{\small 71.49}  & \textbf{\small 71.19} & \underline{\small 63.78} & \underline{\small 63.29}  & {\small 63.24} & {\small 62.49}\\
{\small \taughtpl-div-k} & {\small 63.13} & {\small 63.18} &  {\small 60.77} & {\small 59.85} & {\small 70.00} & \underline{\small 69.89} & {\small 62.33} & {\small 62.27} & \textbf{\small 63.88} & {\small 61.86}\\

\hline
\hline
{\small Delta} & {\small 2.45} & {\small 2.59} & {\small 1.82} & {\small 5.84} & {\small 1.39} & {\small 0.99} & {\small -0.14} & {\small -0.52} & {\small 0.54} & {\small -0.70}\\
\hline
\end{tabular}
\end{adjustbox}
\caption{\label{app:maintable-zs} This table shows the zero-shot accuracies. The top three rows represent the baselines. The highest accuracy is shown in bold; the second highest is underlined. Delta represents the difference between the best-
performing technique and the best-performing baseline}
\end{table*}

\begin{table*}[t!]
\centering
\begin{adjustbox}{max width=\textwidth}
\begin{tabular}{|p{0.1\linewidth}|p{0.05\linewidth}| p{0.05\linewidth}|p{0.05\linewidth}|p{0.05\linewidth}|p{0.05\linewidth}|p{0.05\linewidth}|p{0.05\linewidth}|p{0.05\linewidth}|p{0.05\linewidth}|p{0.05\linewidth}|}
\hline 

\multicolumn{1}{|c|}{\small -} & \multicolumn{1}{|c|}{\small \genonly} & \multicolumn{1}{|c|}{\small \srconly}& \multicolumn{1}{|c|}{\small\srcgen} &  \multicolumn{1}{|c|}{\small \taught-top-k} &  \multicolumn{1}{|c|}{\small \taught-rand-k}  &  \multicolumn{1}{|c|}{\small \taught-div-k }   &  \multicolumn{1}{|c|}{\small \taughtpl-top-k } &  \multicolumn{1}{|c|}{\small \taughtpl-rand-k } &  \multicolumn{1}{|c|}{\small \taughtpl-div-k } &  \multicolumn{1}{|c|}{\small Delta }\\
\hline
\small dev & \small 51.62 & \small 53.02   & \small 55.05 &  \small 52.42 &  \small 50.44 & \small 52.62  &  \small 48.18 &  \small 49.09  &  \small 48.26 & \small -2.43\\
\hline
\end{tabular}
\end{adjustbox}
\caption{\label{app:gluecos-tablett} Translate-train numbers for GLUECoS Sentiment Analysis }
\end{table*}

\begin{table}[t!]
\centering
\begin{adjustbox}{max width=\textwidth}
\begin{tabular}{|p{0.18\linewidth}|p{0.06\linewidth} p{0.07\linewidth}|p{0.08\linewidth}p{0.08\linewidth}|p{0.06\linewidth}p{0.08\linewidth}|}
\hline 
&\multicolumn{2}{|c|}{\small XNLI Hi}& \multicolumn{2}{|c|}{\small XNLI Ur} & \multicolumn{2}{|c|}{\small XNLI Sw}\\
\hline
 & {\small dev} & \small test & \small dev & \small test & \small dev & \small test \\
\hline
{\small \srcgen} &{\small 70.10} & {\small 70.20} & \textbf{\small 63.92} & \textbf{\small 63.81} & {\small 63.34} & \textbf{\small 63.64}\\
\hline
{\small \taught-amb-k} &  {\small 68.68} & {\small 67.47} & {\small 58.28} & {\small 57.19} & {\small 59.32} & {\small 56.55} \\
{\small \taught-easy-k} & {\small 66.43} & {\small 65.20} & {\small 57.95} & {\small 56.38} & {\small 59.28} & {\small 57.05} \\
{\small \taughtpl-amb-k} &  \textbf{\small 72.53} & \textbf{\small 71.67} & \underline{\small 63.90} & \underline{\small 63.58} & \textbf{\small 63.86} & \underline{\small 63.25} \\
{\small \taughtpl-easy-k} &  {\small 69.68} & {\small 69.21} & {\small 62.67} & {\small 61.22} & {\small 61.59} & {\small 61.49}  \\
\hline
{\small Delta} & {\small 2.43} & {\small 1.47} & {\small -0.02} & {\small -0.23} & {\small 0.52} & {\small -0.39} \\
\hline
\end{tabular}
\end{adjustbox}
\caption{\label{app:ambi-easy-xnlizs} Zero-shot accuracies using amb-k and easy-k selection (see section~\ref{dataselect}). Delta is (best score  - \srcgen) score.}
\end{table}

\section{Zero-shot evaluations} \label{app:zsresults}
Tables \ref{app:maintable-zs}, \ref{app:ambi-easy-xnlizs}, \ref{app:hardsoft-tbizs}, \ref{app:randomsubsetzs} shows the zero-shot results of the baselines and our techniques. In table \ref{app:maintable-zs}, except for a few languages, our techniques beat the baselines by a reasonable amount. Table \ref{app:gluecos-tablett} shows the translate-train numbers of the GLUECos task. The evaluation set of GLUECos is in romanized Hindi-English (code-switched text), we suspect this to be the reason for the negative delta.

\section{Code-mixed Text generation}
\label{sec:codemixed}
To generate code-mixed text in mixed-script format for the GLUECos sentiment analysis task, we trained a mt0-xl \citep{mt0} model. The pre-trained model weights were downloaded from the HuggingFace repo\footnote{https://huggingface.co/bigscience/mt0-xl}. The training is done using the huggingface's Trainer API. We made use of Lora \citep{lora} PEFT technique \citep{peft} to train the model, with the hyperparameters: alpha=32, dropout=0.05, r=16, and Lora matrices being applied to query, key, and value attention matrices. The model is trained for 15 epochs, train batch size of 16, gradient accumulation steps of 4, learning rate of 2e-4, max grad norm of 0.3, and warmup ratio of 0.03. The best model checkpoint was selected using the evaluation loss. The data for training the model was obtained from \citep{cocoa}. Table \ref{code-mixed-text} shows few examples. Before generating the translate-train code-mixed text, we first translate the SST5 English train set to Hindi, which we then feed into our model to get the Hinglish mixed-script outputs.

\begin{table}[t!]
\centering
\begin{adjustbox}{max width=\textwidth}
\begin{tabular}{|p{0.15\linewidth}|p{0.08\linewidth}|p{0.08\linewidth}|p{0.08\linewidth}|p{0.08\linewidth}|}
\hline 
&\multicolumn{2}{c}{\small Marsentiment}& \multicolumn{2}{|c|}{\small XNLI Hi}  \\
\hline
&\multicolumn{2}{c}{\small \taught-top-k} & \multicolumn{2}{|c|}{\small \taught-top-k}\\
\hline 	
 &\multicolumn{1}{c}{\small dev}& \multicolumn{1}{|c|}{\small test} &\multicolumn{1}{c}{\small dev}& \multicolumn{1}{|c|}{\small test}\\
\hline
{\small CE} & {\small 33.72} & {\small 33.65} & {\small 43.70} & {\small 43.28}\\
\hline 
{\small KLD} & \textbf{\small 66.34} & \textbf{\small 66.54} & \textbf{\small 70.82} & \textbf{\small 69.81}\\
\hline
\hline
{\small Delta} & \small 32.62 & \small 32.89 & \small 27.12 & \small 26.53\\
\hline
\end{tabular}
\end{adjustbox}
\caption{ Compare training of \taught-top-k model with teacher-soft vs teacher-hard labels. Delta represents the difference between the accuracy (zero-shot) for the student trained via soft labels and the student trained using hard labels.}
\label{app:hardsoft-tbizs}
\end{table}

\begin{table}[t!]
\centering
\begin{adjustbox}{max width=\textwidth}
\begin{tabular}{|p{0.3\linewidth}|p{0.09\linewidth}|p{0.09\linewidth}|}
\hline 
&\multicolumn{2}{c|}{\small \taught-top-k} \\
\hline
&\multicolumn{1}{c}{\small dev}& \multicolumn{1}{|c|}{\small test}\\
\hline
{\small 2500 pos, 2500 neg, 2500 neu} & {\small 66.34} & \textbf{\small 66.54}\\
\hline 
{\small 3000 pos, 3000 neg, 1500 neu} & {\small 64.20} & {\small 63.94}\\
\hline
{\small 3500 pos, 3500 neg, 500 neu} & {\small 65.59} & {\small 65.11}\\
\hline
{\small 3750 pos, 3750 neg, 0 neu} & \textbf{\small 66.75} & {\small 66.18}\\
\hline
\end{tabular}
\end{adjustbox}
\caption{Probing the effect of class imbalance on Marathi sentiment task. pos, neg, and neu indicate positive, negative, and neutral classes.  The best numbers (zero-shot) in a column are highlighted.}
\label{app:randomsubsetzs}
\end{table}

\section{LLAMA-2 Generation Details}
\label{sec:llama2gen}
We use LLAMA-2 13b-chat-hf model for all our generations, as it is a recent state-of-the-art and open-source model. We use a 4-bit quantized version of the model owing to memory constraints. For quantization, we make use of the Bitsandbytes library \citep{bitsandbytes}. We use nucleus sampling for all our generations with p=0.9, we avoid repeating bi-grams, and we keep a temperature between 1.5-2.5 depending on the response of the model to the input prompt. We generated approximately 2lac sentences for each task, which came down to ~1.5lac sentences post-processing and cleaning. We took a random subset of 1.3 lac sentences from the total generations, to which we then applied data selection techniques.

\section{LLAMA-2 Prompt Details}
\label{sec:llama2prompt}
Tables \ref{hiproduct-prompts}, \ref{marsentimentpolitical-prompts}, \ref{marsentimentst-prompts}, \ref{marsentimentgeneric-prompts}, \ref{marsentimentmovie-prompts}, \ref{xnlipremises-prompts}, \ref{xnlihypo-prompts}, \ref{lawdomain-prompts}, \ref{medicaldomain-prompts} show the prompts we used to generate data for different target tasks.%
\footnote{We also tried generating task-specific sentences without specifying any domain information. This resulted in noisier generations compared to when we added a domain description in the prompt. This might be an artefact specific to the LLM we used, and needs further investigation.}
Custom prompts are employed for each target class. In the case of the Hindi product, where the goal is to generate product-specific reviews, numerous neutral (class-conditioned) generations still exhibited subtle indications of positive or negative sentiments. Consequently, we generated additional neutral sentences using a slightly modified prompt to achieve a balanced distribution of classes in the augmented dataset.

For the Marathi Sentiment target task, we devise specific prompts for various domains such as political tweets, generic tweets, subtitles, and movie reviews. To create premise-hypothesis pairs for XNLI, we initially generate the premises providing domain information in the prompts. The hypothesis is then generated using the respective premise as input along with the NLI label (entailment, neutral, and contradiction). Tables \ref{diffsentimentdomains}, \ref{diffnlidomains} show the generations belonging to different domains for sentiment classification and natural language inference (NLI) task.

\section{Computational Budget}
Training a student model on an Nvidia A100 GPU with 80G of RAM took \textasciitilde{100} mins for 15 epochs. We utilized the same GPUs for text generation. Generating \textasciitilde{2} lac sentences of max-length 256 tokens using LLAMA-2 13B model with a batch size of 60 took \textasciitilde{20} hrs using a single GPU and \textasciitilde{50} GB of RAM.

\section{Annotator Details}\label{annotator-details}
Both annotators had professional competence in English. The instructions given for the two tasks are listed below:
\begin{enumerate}
\item Sentiment Task: Given a sentence, identify the polarity of sentiment which could be one of the three types: positive, negative and neutral.
\item NLI Task: Given two pieces of text called premise and hypothesis, mark the pair as “entailment” if premise entails the hypothesis, “contradiction” if premise contradicts the hypothesis and “neutral” if there’s neither entailment nor contradiction i.e the factuality of two statements is independent from each other.
\end{enumerate}

\begin{table*}[t!]
\centering
\begin{tabular}{|p{0.32\linewidth}|p{0.32\linewidth}|}
\hline
 \small\textbf{Input text} & \small\textbf{Code-mixed text} \\
 \hline
\small {Ye baccho ko le jane layak hai.} & \small {Ye \emph{children} ko le jane layak hai.} \\
\hline
\small {vinsent gailo is phraanseesee shokar mein ghar par apne saamaany bure ladake kee ajeeb bhoomika nibha raha hai.} & \small {\emph{Vincent Galo} is \emph{French} shokar mein ghar par apne \emph{normal bad boy} ki \emph{odd role} nibha raha hai.} \\
\hline
\end{tabular}
\caption{\label{code-mixed-text}Examples of code-mixed generation by our trained model. The above examples are transliterated for ease of reading, the words which are translated to English by the model are emphasized.
}
\end{table*}

\begin{table*}[t!]
\centering
\begin{adjustbox}{max width=\textwidth}
\begin{tabular}{|p{0.23\linewidth}|p{0.23\linewidth}|p{0.23\linewidth}|p{0.23\linewidth}|}
\hline
 \small\textbf{Positive} & \small\textbf{Negative} & \small \textbf{Neutral} & \small \textbf{Neutral Add.}\\
 \hline
\small {<s>[INST] <<SYS>>
    You are a user providing reviews on travels, movies and various electronic gadgets. Please only generate the review without any additional content before or after.
    <</SYS>>
    
    Please generate a single review in not more than two short sentences on one of the system specified products/movies/travels indicating a positive sentiment.[/INST]} & \small {<s>[INST] <<SYS>>
    You are a user providing reviews on travels, movies and various electronic gadgets. Please only generate the review without any additional content before or after.
    <</SYS>>
    
    Please generate a single review in not more than two short sentences on one of the system specified products/movies/travels indicating a negative sentiment.[/INST]} & \small {<s>[INST] <<SYS>>
    You are a user who talks about travels, movies and various electronic gadgets in a fact-based, and non-opinionated manner. Please don't involve emotional language or bias. Please only generate the description without any additional content before or after.
    <</SYS>>
    
    Please generate a single and very short sentence on one of the system specified products/movies/travels. It should provide very general information.[/INST]} & \small {<s>[INST] <<SYS>>
    You are a user who talks about travels, movies and various electronic gadgets in a fact-based, and non-opinionated manner. Please don't involve emotional language or bias. Please only generate the description without any additional content before or after.
    <</SYS>>
    
    Please generate a single and very short sentence on one of the system specified products/movies/travels. It should provide very general information, strictly do not use positive words or adjectives.[/INST]}\\
\hline
\end{tabular}
\end{adjustbox}
\caption{\label{hiproduct-prompts}Prompts for HinProduct Task 
}
\end{table*}

\begin{table*}[t!]
\centering
\begin{adjustbox}{max width=\textwidth}
\begin{tabular}
{|p{0.32\linewidth}|p{0.32\linewidth}|p{0.32\linewidth}|}
\hline
 \small\textbf{Positive} & \small\textbf{Negative} & \small \textbf{Neutral}\\
 \hline
\small {<s>[INST] <<SYS>>
    You are a user, political figure or activist who tweets a variety of thoughts and perspectives on current affairs. Please only generate the tweet without any additional content before or after.
    <</SYS>>
    
    Please generate a single tweet in not more than two sentences indicating a positive sentiment with no hashtags, and minimal noise.[/INST]} & \small {<s>[INST] <<SYS>>
    You are a user, political figure or activist who tweets a variety of thoughts and perspectives on current affairs. Please only generate the tweet without any additional content before or after.
    <</SYS>>
    
    Please generate a single tweet in not more than two sentences indicating a negative sentiment with no hashtags, and minimal noise.[/INST]} & \small {<s>[INST] <<SYS>>
    You are a user, political figure or activist who tweets a variety of thoughts and perspectives on current affairs in a fact-based, and non-opinionated manner. Please don't involve emotional language or bias. Please only generate the tweet without any additional content before or after.
    <</SYS>>
    
    Please generate a single tweet in not more than two sentences with no hashtags, and minimal noise. It should provide very general information.[/INST]}\\
\hline
\end{tabular}
\end{adjustbox}
\caption{\label{marsentimentpolitical-prompts}Prompts for MarSentiment Task, specific to political tweets
}
\end{table*}

\begin{table*}[t!]
\centering
\begin{adjustbox}{max width=\textwidth}
\begin{tabular}{|p{0.32\linewidth}|p{0.32\linewidth}|p{0.32\linewidth}|}
\hline
 \small\textbf{Positive} & \small\textbf{Negative} & \small \textbf{Neutral}\\
 \hline
\small {<s>[INST] <<SYS>>
    Please generate subtitles from any situational comedy TV show. Please only generate the subtitle without any additional content before or after.
    <</SYS>>
    
    Please only generate a single sentence taken from the subtitles indicating a positive sentiment without specifying the speaker or any other plot details.[/INST]} & \small {<s>[INST] <<SYS>>
    Please generate subtitles from any situational comedy TV show. Please only generate the subtitle without any additional content before or after.
    <</SYS>>
    
    Please only generate a single sentence taken from the subtitles indicating a negative sentiment without specifying the speaker or any other plot details.[/INST]} & \small {<s>[INST] <<SYS>>
    Please generate subtitles from any situational comedy TV show in a fact-based, and non-opinionated manner. Please don't involve emotional language or bias. Please only generate the subtitle without any additional content before or after.
    <</SYS>>
    
    Please only generate a single sentence taken from the subtitles without specifying the speaker or any other plot details. It should provide very general information.[/INST]}\\
\hline
\end{tabular}
\end{adjustbox}
\caption{\label{marsentimentst-prompts}Prompts for MarSentiment Task, specific to subtitles
}
\end{table*}

\begin{table*}[t!]
\centering
\begin{adjustbox}{max width=\textwidth}
\begin{tabular}{|p{0.32\linewidth}|p{0.32\linewidth}|p{0.32\linewidth}|}
\hline
 \small\textbf{Positive} & \small\textbf{Negative} & \small \textbf{Neutral}\\
 \hline
\small {<s>[INST] <<SYS>>
    You are a user who tweets on a variety of domains. Please only generate the tweet without any additional content before or after.
    <</SYS>>
    
    Please generate a single tweet in not more than two sentences indicating a positive sentiment with no hashtags, and minimal noise.[/INST]} & \small {<s>[INST] <<SYS>>
    You are a user who tweets on a variety of domains. Please only generate the tweet without any additional content before or after.
    <</SYS>>
    
    Please generate a single tweet in not more than two sentences indicating a negative sentiment with no hashtags, and minimal noise.[/INST]} & \small {<s>[INST] <<SYS>>
    You are a user who tweets on a variety of domains in a fact-based, and non-opinionated manner. Please don't involve emotional language or bias. Please only generate the tweet without any additional content before or after.
    <</SYS>>
    
    Please generate a single tweet in not more than two sentences with no hashtags, and minimal noise. It should provide very general information.[/INST]}\\
\hline
\end{tabular}
\end{adjustbox}
\caption{\label{marsentimentgeneric-prompts}Prompts for MarSentiment Task, specific to generic tweets
}
\end{table*}

\begin{table*}[t!]
\centering
\begin{adjustbox}{max width=\textwidth}
\begin{tabular}{|p{0.32\linewidth}|p{0.32\linewidth}|p{0.32\linewidth}|}
\hline
 \small\textbf{Positive} & \small\textbf{Negative} & \small \textbf{Neutral}\\
 \hline
\small {<s>[INST] <<SYS>>
    You are a user who provides reviews on a variety of Indian movies. Please only generate the review without any additional content before or after.
    <</SYS>>
    
    Please generate a single review in not more than two sentences indicating a positive sentiment with minimal noise.[/INST]} & \small {<s>[INST] <<SYS>>
    You are a user who provides reviews on a variety of Indian movies. Please only generate the review without any additional content before or after.
    <</SYS>>
    
    Please generate a single very short review in not more than two sentences indicating a negative sentiment with minimal noise.[/INST]} & \small {<s>[INST] <<SYS>>
    You are a user who provides reviews on a variety of Indian movies in a fact-based, and non-opinionated manner. Please don't involve emotional language or bias. Please only generate the review without any additional content before or after.
    <</SYS>>
    
    Please generate a single, short review in not more than two sentences with minimal noise. It should provide very general information.[/INST]}\\
\hline
\end{tabular}
\end{adjustbox}
\caption{\label{marsentimentmovie-prompts}Prompts for MarSentiment Task, specific to movie reviews
}
\end{table*}

\begin{table*}[t!]
\centering
\begin{adjustbox}{max width=\textwidth}
\begin{tabular}{|p{0.32\linewidth}|p{0.32\linewidth}|p{0.32\linewidth}|}
\hline
 \small\textbf{Travel} & \small\textbf{Government} & \small \textbf{Fiction}\\
 \hline
\small {<s>[INST] <<SYS>>You are a user who talks about other people's traveling experiences. Please only generate the traveling experience in a single sentence without any additional content before or after. <</SYS>>Please generate a single and short sentence belonging to the travel domain.[/INST]} & \small {<s>[INST] <<SYS>>Your job is to generate a diverse sentence in the domain provided by the user. Please only generate the sentence without any additional content before or after. <</SYS>>Please generate a single and short sentence belonging to the government domain. [/INST]} & \small {<s>[INST] <<SYS>>Your job is to generate a diverse sentence in the domain provided by the user. Please only generate the sentence without any additional content before or after. <</SYS>>Please generate a single and short sentence belonging to the fiction domain. [/INST]}\\
\hline
\end{tabular}
\end{adjustbox}
\caption{\label{xnlipremises-prompts}Prompts for XNLI premises.
}
\end{table*}

\begin{table*}[t!]
\centering
\begin{adjustbox}{max width=\textwidth}
\begin{tabular}{|p{0.32\linewidth}|p{0.32\linewidth}|p{0.32\linewidth}|}
\hline
 \small\textbf{Entailment} & \small\textbf{Neutral} & \small \textbf{Contradiction}\\
 \hline
\small {<s>[INST] <<SYS>>Please generate a single sentence that is implied from the sentence provided by the user. The sentence generated could encompass either a portion or the entirety of the information contained in the given sentence. Please ensure the generations are grammatically correct. Please only share the generation without any additional content before or after.<</SYS>>Sentence: {}[/INST]} & \small {<s>[INST] <<SYS>>Please generate a single sentence related to the user provided sentence that is neither entailed nor contradicts the user provided sentence. Please ensure the generations are grammatically correct. Please only share the generation without any additional content before or after.<</SYS>>Sentence: {}[/INST]} & \small {<s>[INST] <<SYS>>Please generate a single sentence that logically contradicts the information provided in the sentence given by the user. Please ensure the generations are grammatically correct. Please only share the generation without any additional content before or after.<</SYS>>Sentence: {}[/INST]}\\
\hline
\end{tabular}
\end{adjustbox}
\caption{\label{xnlihypo-prompts}Prompts for XNLI hypothesis, \{\} is replaced with the premise for which a hypothesis needs to be generated
}
\end{table*}

\begin{table*}[t!]
\centering
\begin{adjustbox}{max width=\textwidth}
\begin{tabular}{|p{0.32\linewidth}|p{0.32\linewidth}|p{0.32\linewidth}|}
\hline
 \small\textbf{Positive} & \small\textbf{Negative} & \small \textbf{Neutral}\\
 \hline
\small {<s>[INST] <<SYS>>
    You are a lawyer who talks about laws. Please only generate the sentence without any additional content before or after.
    <</SYS>>
    
    Please generate a single sentence indicating a positive sentiment with minimal noise.[/INST]} & \small {<s>[INST] <<SYS>>
    You are a lawyer who talks about laws. Please only generate the sentence without any additional content before or after.
    <</SYS>>
    
    Please generate a single sentence indicating a negative sentiment with minimal noise.[/INST]} & \small {<s>[INST] <<SYS>>
    You are a lawyer who talks about laws in a fact-based, and non-opinionated manner. Please don't involve emotional language or bias. Please only generate the sentence without any additional content before or after.
    <</SYS>>
    
    Please generate a single sentence with minimal noise. It should provide very general information.[/INST]}\\
\hline
\end{tabular}
\end{adjustbox}
\caption{\label{lawdomain-prompts}Prompts for Law domain
}
\end{table*}

\begin{table*}[t!]
\centering
\begin{adjustbox}{max width=\textwidth}
\begin{tabular}{|p{0.32\linewidth}|p{0.32\linewidth}|p{0.32\linewidth}|}
\hline
 \small\textbf{Positive} & \small\textbf{Negative} & \small \textbf{Neutral}\\
 \hline
\small {<s>[INST] <<SYS>>
    You are a doctor who talks about medicine. Please only generate the sentence without any additional content before or after.
    <</SYS>>
    
    Please generate a single sentence indicating a positive sentiment with minimal noise.[/INST]} & \small {<s>[INST] <<SYS>>
    You are a doctor who talks about medicine. Please only generate the sentence without any additional content before or after.
    <</SYS>>
    
    Please generate a single sentence indicating a negative sentiment with minimal noise.[/INST]} & \small {<s>[INST] <<SYS>>
    You are a doctor who talks about medicine in a fact-based, and non-opinionated manner. Please don't involve emotional language or bias. Please only generate the sentence without any additional content before or after.
    <</SYS>>
    
    Please generate a single sentence with minimal noise. It should provide very general information.[/INST]}\\
\hline
\end{tabular}
\end{adjustbox}
\caption{\label{medicaldomain-prompts}Prompts for medical domain
}
\end{table*}

\begin{table*}[t!]
\centering
\begin{adjustbox}{max width=\textwidth}
\begin{tabular}{|p{0.2\linewidth}|p{0.23\linewidth}|p{0.23\linewidth}|p{0.23\linewidth}|}
\hline
\small\textbf{Target domain} & \small\textbf{Positive} & \small\textbf{Negative} & \small \textbf{Neutral}\\
 \hline
\small Marsentiment domain & \small The acting and performances in this movie are truly outstanding. The story is engaging and the script is witty, making for a thoroughly entertaining watch. & \small This project is doomed, and I have no idea how we're going to pull it off! & \small They think I built a spaceship for my thesis? I wish!\\
\hline
\small Law sentiment domain & \small The new legislation promotes community engagement and supports sustainable development, demonstrating a thoughtful approach to community growth and preservation. & \small The current state of the law in this matter is incredibly disappointing and leaves much to be desired. & \small Mandated paid sick leave is regulated by The District of Columbia under DC Law Title 32.\\
\hline
\small Medical sentiment domain & \small This latest breakthrough in cancer treatment is truly inspiring, and offers new hope for patients. & \small I'm concerned about the rising number of adverse reactions to this new drug - it's not worth the risk. & \small Doctors today often perform robot-assisted spinal fusion surgeries for degenerative disc conditions!\\
\hline
\end{tabular}
\end{adjustbox}
\caption{\label{diffsentimentdomains}Generations for different sentiment domains.
}
\end{table*}

\begin{table*}[t!]
\
\begin{adjustbox}{max width=\textwidth}
\begin{tabular}{|p{0.2\linewidth}|p{0.23\linewidth}|p{0.23\linewidth}|p{0.23\linewidth}|}
\hline
\small\textbf{Target domain} & \small\textbf{Entailment} & \small\textbf{Contradiction} & \small \textbf{Neutral}\\
 \hline
\small MNLI Domain & \small The government is working to provide affordable healthcare to all citizens.; Government efforts are being made to offer health care that citizens can afford. & \small Karla hiked through the misty hills of Iceland, camera in hand and spirit for adventure.; Karla stayed at home, enjoying a relaxing day off from hikes and adventures. & \small Climbed to the summit of Mount Kilimanjaro at dawn and saw the breathtaking sunrise over the plains below.; The group celebrated their triumph with a picture at the peak while watching the stunning dawn.\\
\hline
\small Law NLI domain & \small The judge granted a rare form of temporary relief to the small business owner, allowing her to retain possession of her commercial property pending a full trial.; The judge provided temporary possession relief, permitting the small enterprise owner to keep her property for the trial's duration. & \small The Supreme Court struck down the contested law, deeming it unconstitutional and a violation of individual privacy rights.; The contended law was upheld by the supreme court, declaring it Constitutional as it adequately protects individual freedoms and promotes public safety. & \small The court found the defendant guilty of wire fraud and sentenced them to 250 hours of community service.; The defendent will serve their community servie in an independent living center for the elderly.\\
\hline
\small Medical NLI domain & \small Novel Biomarkers Found to Diagnose Acute Heart Failure in High-Risk Patients.; heart failure can be diagnosed using novel biomakers in high-risk patients. & \small The patient's CT scan revealed a previously undetected tumor in her lung.; The patient had no evidence of any lung tumors on her CT Scan. & \small The novel antiviral drug successfully treated the patient's rare and aggressive strain of influenza.; The doctor praised the effectiveness of the antibiotic in treating the elderly patient.\\
\hline
\end{tabular}
\end{adjustbox}
\caption{\label{diffnlidomains}Generations for different NLI domains. The premise and hypothesis in each column are separated by ';'.
}
\end{table*}
\end{document}